\pdfoutput=1
\documentclass[11pt]{article}

\usepackage{ACL2023}
\usepackage{multirow}
\usepackage{graphicx}
\usepackage{times}
\usepackage{latexsym}

\usepackage[T1]{fontenc}
\usepackage[utf8]{inputenc}

\usepackage{microtype}

\usepackage{inconsolata}

\title{FreCDo: A Large Corpus for French Cross-Domain Dialect Identification}

\author{Mihaela G\u{a}man\\
  University of Bucharest, Romania\\\And
  Adrian-Gabriel Chifu \\
  Aix-Marseille Universit\'{e}, France\\\AND
  William Domingues \\
  Aix-Marseille Universit\'{e}, France\\\And
  Radu Tudor Ionescu\thanks{\hspace{0.2cm}Corresponding author: \texttt{raducu.ionescu@gmail.com}}\\
  University of Bucharest, Romania}

\begin{document}
\maketitle
\begin{abstract}
\vspace{-0.2cm}
We present a novel corpus for French dialect identification comprising 413,522 French text samples collected from public news websites in Belgium, Canada, France and Switzerland. To ensure an accurate estimation of the dialect identification performance of models, we designed the corpus to eliminate potential biases related to topic, writing style, and publication source. More precisely, the training, validation and test splits are collected from different news websites, while searching for different keywords (topics). This leads to a \textbf{Fre}nch \textbf{c}ross-\textbf{do}main (FreCDo) dialect identification task. We conduct experiments with four competitive baselines, a fine-tuned CamemBERT model, an XGBoost based on fine-tuned CamemBERT features, a Support Vector Machines (SVM) classifier based on fine-tuned CamemBERT features, and an SVM based on word n-grams. Aside from presenting quantitative results, we also make an analysis of the most discriminative features learned by CamemBERT. Our corpus is available at \url{https://github.com/MihaelaGaman/FreCDo}.
\end{abstract}

\section{Introduction}
\vspace{-0.15cm}
Dialect identification is an actively studied topic in computational linguistics, one of the most significant drivers for research in this area being the yearly shared tasks on dialect identification organized at VarDial \cite{Chakravarthi-VarDial-2021,Gaman-VarDial-2020b,Malmasi-VarDial-2016,Zampieri-VarDial-2017,Zampieri-VarDial-2018,Zampieri-VarDial-2019}. Identifying dialects is important for analyzing and understanding the evolution of languages across time and geographical regions. The task is also useful in forensics analysis, where the dialect of a text sample might reveal some information about the author of the respective text. Despite the valuable applications, French dialect identification is not extensively studied in literature \cite{blondeau2021,jakubivcek2013,Tan2014}. Since French is spoken across multiple countries across the globe, we believe that French dialect identification should be given more attention. To this end, we present a novel and large dialect identification corpus composed of over 400K French text samples representing four dialects: Belgian French, Canadian French, Hexagonal French and Swiss French. 

Due to the high number of data samples, our corpus is suitable for the training and evaluation of modern deep language models, such as CamemBERT \cite{Martin-ACL-2020}. However, machine learning models are prone to overfitting the biases of the training data. To ensure a fair estimation of the generalization capacity of such models, we designed the corpus to mitigate potential biases related to topic, writing style, and publication source, which could lead to falsely increasing the accuracy of the evaluated models. This is achieved by separating the text samples into training, validation and test sets, such that the examples in each of the three splits are from different news websites than the other two splits. Similarly, the data samples in each split are collected from the results retrieved while searching for a set of keywords (topics) that are distinct from the other splits. Our data collection process leads to a \textbf{Fre}nch \textbf{c}ross-\textbf{do}main (FreCDo) dialect identification task.

We conduct experiments with four baselines, namely a fine-tuned CamemBERT model, an XGBoost \cite{Chen-SIGKDD-2016} based on fine-tuned CamemBERT features, a Support Vector Machines (SVM) \cite{Cortes-ML-1995} classifier based on fine-tuned CamemBERT features, and an SVM based on word n-grams. Although our baselines are fairly competitive, the empirical results indicate that the French cross-domain dialect identification task is very challenging, the highest macro $F_1$ score being $39.67\%$. To determine if the top scoring model, the standalone CamemBERT, learned training set biases, we analyze the most discriminative features of the respective model.

In summary, our contribution is threefold:
\begin{itemize}
    \item \vspace{-0.25cm} We introduce a novel large-scale corpus for French cross-domain dialect identification.
    \item \vspace{-0.25cm} We train and evaluate four competitive baselines, three of them being based on a modern deep language model based on the transformer architecture, i.e.~CamemBERT.
    \item \vspace{-0.25cm} We analyze the most discriminative features of CamemBERT, revealing the dialectal patterns learned by this state-of-the-art model.
\end{itemize}

\section{Related Work}
\vspace{-0.15cm}

One of the most important research efforts in the area of dialect identification is represented by the DSL Corpus Collection (DLSCC)\footnote{\url{http://ttg.uni-saarland.de/resources/DSLCC/}} \cite{Tan2014}, associated with the VarDial workshop. The corpus is at its fifth iteration and it contains short excerpts of journalistic texts. The latest version (4.0) contains six groups of languages with their corresponding dialects. For instance, the Hexagonal French and Canadian French represent one of the groups. Our data set contains two more French dialects (Swiss and Belgian) and is considerably larger (by an order of magnitude).

CFPR\footnote{\url{https://cfpr.huma-num.fr}} is a regional spoken French corpus containing voice recordings with their transcriptions. The corpus contains 388,123 words (30 hours of recordings) produced by 172 speakers from around the world. Most of the speakers (130) are from Europe, out of which 17 are from Belgium and 6 from Switzerland. There are 5 speakers from Canada, 17 from Northern Africa and 2 from Lebanon. Even though the corpus contains more dialects than our corpus, we observe that the CFPR corpus is very unbalanced. Moreover, it does not take into account the concept of topics, i.e.~samples can also be discriminated based on the discussed topics.

Another considerable French resource is the French Web Corpus (frTenTen)\footnote{\url{https://www.sketchengine.eu/frtenten-french-corpus/\#toggle-id-3}}, a part of the TenTen corpus family \cite{jakubivcek2013}. The latest iteration of this French corpus (frTenTen17, from 2017) contains 5.7 billion words and it was annotated by FreeLing (lemmatization, part-of-speech tagging, etc.). This corpus is not designed specifically for dialect identification, but the dialects could be inferred from the metadata containing the web domain. Of course, this requires some processing efforts. As the CFPR corpus, frTenTen is not focused on cross-topic or cross-source analysis either.
There are other researchers focusing on French variants or dialects, but with different scope. For instance, \newcite{blondeau2021} have studied the Montreal French intrinsic evolution. 

\begin{table*}
\centering
\small{
\begin{tabular}{|l|l|c|l|r|r|}
\hline
\textbf{Split} & \textbf{Topics: FR (EN)} & \textbf{Country} & \textbf{Source} &  \textbf{\#Samples} &  \textbf{\#Tokens} \\
\hline
\hline
\multirow{5}{*}{\textbf{Train}}& & BE & dhnet & 121,746 & 11,619,874 \\
&Guerre (War),                            & CA & radio-canada & 34,003 & 2,505,254 \\
&Ukraine (Ukraine)                             & CH & letemps & 141,261 & 12,719,203 \\
&                            & FR & cnews & 61,777 & 6,397,943 \\
\cline{2-6}
& \multicolumn{3}{|r|}{\textbf{Total:}} & 358,787 & 33,242,274 \\
\hline
\multirow{5}{*}{\textbf{Validation}} & & BE & ln24 & 7,723 & 824,871 \\
& Russie (Russia)                            & CA & journal-metro & 171 & 17,061 \\
& \'{E}tats-Unis (United States)                             & CH & le-courrier & 5,244 & 476,338 \\
&                            & FR & huffpost & 4,864 & 434,547 \\
\cline{2-6}
&    \multicolumn{3}{|r|}{\textbf{Total:}} & 18,002 & 1,752,817 \\
\hline
  \multirow{5}{*}{\textbf{Test}}
& & BE & lavenir & 15,235 & 1,227,263 \\
& R\'{e}chauffement climatique (Global warming)  & CA & science-presse & 944 & 86,724 \\
& Covid (Covid)                             & CH & 24heures & 9,824 & 910,700 \\
&                            & FR & franceinfo & 10,730 & 848,845 \\
\cline{2-6}
&    \multicolumn{3}{|r|}{\textbf{Total:}} & 36,733 & 3,073,532 \\
\hline
\end{tabular}
}
\vspace{-0.2cm}
\caption{\label{topics-country-source}
Our large-scale corpus is composed of more than 400K data samples, containing a total of more than 38M tokens. The corpus is divided into training, validation and test sets, such that the overlap between topics and publication sources across its subsets is reduced as much as possible.\vspace{-0.35cm}
}
\end{table*}

\vspace{-0.1cm}
\section{Corpus}
\vspace{-0.15cm}

The texts comprising the FreCDo corpus have been extracted from news websites from four countries: Belgium (BE), Canada (CA), Switzerland (CH) and France (FR). %Thus, we focus on written (not spoken) French. 
Since the texts are extracted from written news articles, the only connection to spoken French is coming from manual transcriptions of quotes or interviews included in the respective articles. Dialect-wise, written texts (including literary transcriptions of spoken language) are harder to identify than spoken utterances, since professional writing is more formal (standardized).

We have considered several general, country independent topics in order to be able to obtain a large volume of data, while also mitigating possible dialect-specific biases (since generic keywords are to be found in all the sources), and we have queried the news websites with them. The topics are: Guerre (War), Ukraine (Ukraine), Russie (Russia), \'{E}tats-Unis (United States), R\'{e}chauffement climatique (Global warming) and Covid (Covid). The timeframe for data collection is between November 2021 and March 2022, but the articles can be older than the start date. We queried the listed websites (see Table~\ref{topics-country-source}) with the aforementioned keywords and we have crawled the returned pages. For some queries, we obtained a large number of news articles from certain websites. In such cases, we used undersampling to preserve a relative class balance of the splits, randomly selecting articles from the very large retrieved lists, while keeping all returned articles from the other lists.

From each extracted article, we kept the title, the headline and the body of the article. We employed spaCy\footnote{\url{https://spacy.io} (v3.2.2, ``fr\_core\_news\_sm'' model)} to identify named entities and replace them with the special token \$NE\$, in order to avoid any topic or country specific biases. We next split the texts by paragraphs. A paragraph is defined by a group of three consecutive sentences (the sentences being delimited by punctuation). One paragraph represents a data sample (instance) in the data set. Following previous work in building corpora for dialect identification \cite{Butnaru-ACL-2019}, the class labels are based on the website domains. Hence, human intervention was not used in the labeling process. The considered topics and their associated publication sources are summarized in Table~\ref{topics-country-source}. We underline that the overlap between topics and sources across the training, validation and test splits is close to marginal. Hence, models learning topical cues, author style or publication source patterns will not benefit from such biases when the evaluation is performed on the designated test set.

Out of the 413,522 samples, our data split leaves the training set with 358,787 samples (86.8\%), and the validation and test sets with 18,002 (4.3\%) and 36,733 (8.9\%) samples, respectively. We notice that the CA dialect is significantly underrepresented in each of the three splits. We did not find sufficient public news articles from Canadian sources (most of the news articles in Canada are only accessible through subscription-based accounts), which explains the small number of samples that were collected from this country. %This also affects our metrics for the CA class, as can be observed from Section 4.2.

\vspace{-0.1cm}
\section{Experiments}
\vspace{-0.1cm}
\subsection{Baseline Models}
\vspace{-0.1cm}

We publicly release the code to reproduce all baseline models along with the data set.

\noindent
\textbf{Fine-tuned CamemBERT.} The first baseline is a fine-tuned CamemBERT model \cite{Martin-ACL-2020}, which is based on the RoBERTa architecture \cite{Liu-2019}. In the original paper, \citet{Martin-ACL-2020} reported state-of-the-art results on a variety of French downstream tasks, hence our preference for using this architecture as baseline.

Our fine-tuning procedure requires us to encode each textual input into a list of token IDs. For this purpose, we use the CamemBERT tokenizer. Each token is translated into its corresponding 768-dimensional embedding vector. We then add a global average pooling layer to the current architecture, obtaining a Continuous Bag-of-Words (CBOW) representation for each text sequence. Then, we add a Softmax classification layer with four neural units, corresponding to each of the four French dialects targeted in this work, namely Belgium (BE), Canadian (CA), Swiss (CH) and French (FR). Given a text input, the final dialect label is obtained by applying \emph{argmax} on the four probabilities. We fine-tune the model described above for 11 epochs (based on early stopping) on mini-batches of 32 samples, using the Adam with decoupled weight decay (AdamW) optimizer \cite{Loshchilov-ICLR-2019}, with a learning rate of $5\cdot10^{-5}$ and an $\epsilon$ of $10^{-8}$. We tuned the learning rate between $10^{-5}$ and $10^{-4}$ and tested two loss options, cross-entropy vs. negative log-likelihood. All models are trained on a GeForce GTX 1080Ti GPU.

\noindent
\textbf{XGBoost + fine-tuned CamemBERT features.} % Gradient tree boosting \cite{Friedman-2001} is a Machine Learning technique based on training tree ensemble models in an additive manner. % Gradient tree boosting has proved to be effective in solving classification problems \cite{Li-UAI-2010}. Moreover, the method in question is  the ensemble approach of choice in real-world pipelines that run in production \cite{He-DMOA-2014}.
XGBoost \cite{Chen-SIGKDD-2016} is a parallel gradient boosting approach based on iteratively combining decision trees into a single model. %This type of model is focused on solving large-scale tasks with limited computational resources. %At the same time, this method uses shrinkage to address overfitting by reducing the influence of each tree. Column subsampling - borrowed from Random Forests \cite{Breiman-ML-2001} is employed as well and it brings the advantage of computation speed up. 
We apply XGBoost over fine-tuned CamemBERT embeddings. An embedding is a 768-dimensional representation of the \emph{[CLS]} token of each text sample fed into the model. We use $400$ estimators, with a maximum depth of a tree equal to $200$, a learning rate of $7\cdot10^{-2}$, and $\gamma=1$. %The $L_1$ regularization parameter $\alpha$ as well as the $L_2$ regularization term $\lambda$ are both set to $10^{-1}$. 
These hyperparameters are obtained using grid search considering the following intervals: maximum depth between $10$ and $400$, number of estimators between 100 and 400, learning rate between $10^{-2}$ and $2\cdot10^{-1}$, and $\gamma$ between $0$ and $3$. We set the $L_2$ and $L_1$ regularization weights to $\lambda=0.1$ and $\alpha=0.1$, without tuning.

\noindent
\textbf{SVM + fine-tuned CamemBERT features.} We use a Support Vector Machines (SVM) \cite{Cortes-ML-1995} classifier as our third baseline, and apply it on 768-dimensional fine-tuned CamemBERT embeddings. In our experiments, we opt for the linear kernel and the one-vs-rest multi-class classification strategy. We tune the regularization penalty $C$, in a range of values from $10^{-4}$ to $10^{4}$ using grid search. We obtain optimal results with $C=10$. % for the regularization penalty. 

\noindent
\textbf{SVM + word n-grams.} As our fourth baseline, we include an SVM based on word n-grams, a conventional approach commonly used in dialect identification \cite{benites-etal-2019-twistbytes,wu-etal-2019-language}. We tune the n-gram length in the range $1$-$3$, obtaining the best results with word bigrams. As for the third baseline, we also tune the penalty $C$ of the SVM, considering values in the range $10^{-4}$ to $10^{4}$. The optimal value for $C$ is $1$.

\vspace{-0.1cm}
\subsection{Results}
\vspace{-0.1cm}

\begin{table}[!t]
\footnotesize
\setlength{\tabcolsep}{4.0pt}
\begin{center}
 \begin{tabular}{|l | c | c | c | c |} 
 \hline
     &  \multicolumn{2}{|c|}{Validation} & \multicolumn{2}{|c|}{Test}\\
 \cline{2-3}  \cline{4-5}
    Method      & Acc. & Macro  & Acc.  & Macro \\
                &       & F1    &       & F1 \\
 \hline 
 \hline
 Majority class         & 0.4290 & 0.1501 & 0.4147 & 0.1466 \\
 \textbf{CamemBERT}              & \textbf{0.7352} & \textbf{0.4784} & \textbf{0.5584} & \textbf{0.3967} \\ 
 XGBoost + emb          & 0.7280 & 0.4518 & 0.4909 & 0.3477 \\ 
 SVM + emb              & 0.6633 & 0.4534 & 0.4705 & 0.3439 \\ 
 SVM + n-grams          & 0.6560 & 0.4487 & 0.4514 & 0.3108 \\ 
 \hline
\end{tabular}
\end{center}
\vspace{-0.25cm}
\caption{Performance metrics obtained with each of the four baselines versus the majority class baseline, for both validation and test sets.}
\vspace{-0.3cm}
\label{tab_results}
\end{table}

The results reported in Table \ref{tab_results} show that the fine-tuned CamemBERT obtains both the best micro (accuracy) and macro F1 scores in the 4-class French dialect classification task, on both validation and test sets from our corpus. The XGBoost model based on CamemBERT embeddings takes the second place, not too far from the best performing standalone CamemBERT. The SVM based on the fine-tuned embeddings falls further back compared to the other two methods based on deep CamemBERT features, both on the validation, as well as on the test set. The SVM based on word bigrams attains the lowest performance, confirming that the deep CamemBERT features are very useful. We underline that all our baselines obtain better results than the majority (dominant) class baseline. However, the task proves to be significantly challenging, leaving plenty of room for future improvements.

% \begin{figure}[!t]
% \begin{center}
% %\includegraphics[natwidth=1bp,natheight=1bp,width=1.0\linewidth]{categoryBarChart-20.png}
% %\hspace{-0.5cm}
% \includegraphics[width=0.85\linewidth]{bert_preds_val-confusion-mat.jpeg}
% \end{center}
% \vspace*{-0.4cm}
% \caption{Confusion matrix obtained after running the fine-tuned CamemBERT classifier on the validation set. Best viewed in color.}
% \label{cm-val}
% \vspace*{-0.3cm}
% \end{figure}

\begin{figure}[!t]
\begin{center}
%\includegraphics[natwidth=1bp,natheight=1bp,width=1.0\linewidth]{categoryBarChart-20.png}
%\hspace{-0.5cm}
\includegraphics[width=0.89\linewidth]{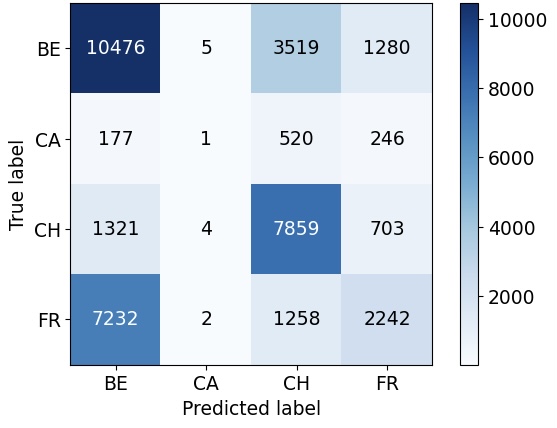}
\end{center}
\vspace*{-0.4cm}
\caption{Confusion matrix obtained after running the fine-tuned CamemBERT classifier on the test set. Best viewed in color.}
\label{cm-test}
\vspace*{-0.3cm}
\end{figure}

%From what we can observe in Fig. \ref{cm-val} as well as in Fig. \ref{cm-test}, 
From what we can observe in Figure \ref{cm-test}, 
our fine-tuned CamemBERT attains more accurate results for the Belgium and Swiss French dialects for which, in their majority, the samples tend to be correctly classified. Less represented in the training set, Canadian French registers the worst results in both evaluation subsets.

\vspace{-0.1cm}
\subsection{Discriminative Feature Analysis}
\vspace{-0.1cm}

For the discriminative feature analysis, we have considered a few correctly classified samples and we have analyzed the features for which CamemBERT has given high scores.
From this manual analysis, we made a few observations. We noticed that a lot of short words and stop words have high CamemBERT scores. Regarding the discriminative terms, for one CH sample, there is the term ``paraphe'' (signature) occurring in the context of a petition signature. In Hexagonal French, one would say ``signature'' (signature) in the aforementioned context. For BE and CH samples, we observed the presence of numerals, such as ``septante'' (70) and ``nonante'' (90), as good indicators for the dialect. In Hexagonal French, the same numerals are based on the addition of 60 and 10 (``soixante-dix'') to get 70, or the multiplication of 4 and 20 followed by an addition with 10 (``quatre-vingt-dix'') to get 90. For CH samples, we also noticed the term ``francs'', which is the Swiss currency. This is quite specific for this country, since France and Belgium have the Euro, and Canada has the Canadian dollar. %For the other dialects we did not find any interesting discriminative features.
In conclusion, there are quite a few noticeable dialectal patterns learned by the model.

\vspace{-0.1cm}
\section{Conclusion}
\vspace{-0.15cm}

In this paper, we presented a novel corpus for cross-domain French dialect identification. We introduced three competitive baselines based on a state-of-the-art language model, CamemBERT, but the reported results showed that the proposed task is very challenging. We believe that a promising future direction for improving performance is to develop an effective approach to avoid overfitting training set biases.

\section*{Limitations}

One important limitation of this article is that the text samples are issued exclusively from news articles. Hence, the variety in terms of text genre is not significant. Another limitation could be the relatively small number of topics that have been considered. Still, the proposed corpus is the largest one for French dialect identification, to date.

Regarding the number of considered dialects, adding more dialects, e.g.~from Northern Africa (Algeria) or Lebanon, would broaden our benchmark. Unfortunately, when we started our data collection process, we did not find sufficient French news samples from other French speaking countries. Thus, perhaps the most important limitation of our data set is the lack of samples representing French dialects from other regions, e.g. Northern Africa.

Considering the large volume of articles included in our corpus, we did not check for aspects such as the true nationality of the authors, which would require intricate manual investigation to find the necessary details about the authors of articles. Indeed, the task of finding the true nationality of the authors can become quite complex, considering that news articles may contain quoted text and that the journalists writing the main piece may publish under a pseudonym, to keep their identity private. To this end, we refrained from performing such investigations.

To replace named entities, we employed an automatic named entity recognizer called spaCy. Since spaCy is not $100\%$ accurate, some named entities were likely not replaced in our corpus. Since we selected different topics for our training, validation and test splits, the remaining named entities are still unlikely to become discriminative features across splits. We thus consider that unchanged named entities represent a minor limitation, requiring no further attention.

%EMNLP 2022 requires all submissions to have a section titled ``Limitations'', for discussing the limitations of the paper as a complement to the discussion of strengths in the main text. This section should occur after the conclusion, but before the references. It will not count towards the page limit.  

%The discussion of limitations is mandatory. Papers without a limitation section will be desk-rejected without review.
%ARR-reviewed papers that did not include ``Limitations'' section in their prior submission, should submit a PDF with such a section together with their EMNLP 2022 submission.

%While we are open to different types of limitations, just mentioning that a set of results have been shown for English only probably does not reflect what we expect. 
%Mentioning that the method works mostly for languages with limited morphology, like English, is a much better alternative.
%In addition, limitations such as low scalability to long text, the requirement of large GPU resources, or other things that inspire crucial further investigation are welcome.

\section*{Ethics Statement}
%Scientific work published at EMNLP 2022 must comply with the \href{https://www.aclweb.org/portal/content/acl-code-ethics}{ACL Ethics Policy}. We encourage all authors to include an explicit ethics statement on the broader impact of the work, or other ethical considerations after the conclusion but before the references. The ethics statement will not count toward the page limit (8 pages for long, 4 pages for short papers).
Our data processing steps ensure anonymity, since the named entities have been replaced with a special token. Hence, names of people have been removed from all samples. Moreover, our data collection process respects the intellectual property of the news articles, since the data samples are formed of three sentence excerpts (paragraphs) from news articles. The data samples are shuffled, so the original news articles cannot be reconstructed. Furthermore, we adhere to the European regulations\footnote{\url{https://eur-lex.europa.eu/eli/dir/2019/790/oj}} that allow researchers to use data in the public web domain for non-commercial research purposes. We will thus release our corpus as open-source under a non-commercial share-alike license agreement.

We acknowledge that spaCy could have missed the names of some people. We will remove samples containing personal details about people, should we receive any specific complaints on this regard via e-mail.

% \section*{Acknowledgements}

\bibliography{anthology,custom}
\bibliographystyle{acl_natbib}

% \appendix

% \section{Example Appendix}
% \label{sec:appendix}

\end{document}